\documentclass{article}
\usepackage[utf8]{inputenc}

\usepackage{microtype}
\usepackage{graphicx}
\usepackage{subfigure}
\usepackage{booktabs} 
\usepackage{hyperref}
\usepackage{amsmath}
\usepackage{amssymb}
\usepackage{todonotes}

\newcommand{\var}[1]{{\operatorname{\textsc{#1}}}}

\usepackage[accepted]{icml2018}

\icmltitlerunning{Counting to Explore and Generalize in Text-based Games}

\newcommand{\ie}{\emph{i.e.},~}
\newcommand{\eg}{\emph{e.g.},~}

\newcommand{\code}[1]{\texttt{#1}}
\newcommand{\literal}[1]{``\texttt{#1}''}

\newcommand{\vocab}{V}

\newcommand{\bc}{\boldsymbol{c}}  

\newcommand{\cmd}[1]{\textbf{\small{\code{#1}}}}

\begin{document}

\twocolumn[
\icmltitle{Counting to Explore and Generalize in Text-based Games}



\icmlsetsymbol{equal}{*}

\begin{icmlauthorlist}
\icmlauthor{Xingdi Yuan}{equal,ms}
\icmlauthor{Marc-Alexandre Côté}{equal,ms}
\icmlauthor{Alessandro Sordoni}{ms}
\icmlauthor{Romain Laroche}{ms}
\icmlauthor{Remi Tachet des Combes}{ms}
\icmlauthor{Matthew Hausknecht}{ms}
\icmlauthor{Adam Trischler}{ms}
\end{icmlauthorlist}

\icmlaffiliation{ms}{Microsoft Research}
\icmlcorrespondingauthor{Eric Yuan}{eric.yuan@microsoft.com}
\icmlcorrespondingauthor{Marc-Alexandre Côté}{macote@microsoft.com}

\icmlkeywords{Machine Learning, Reinforcement Learning, Exploration, Count-based, Text-based Games}

\vskip 0.3in
]


\printAffiliationsAndNotice{\icmlEqualContribution} 

\begin{abstract}
  We propose a recurrent RL agent with an episodic exploration mechanism that helps discovering good policies in text-based game environments. We show promising results on a set of generated text-based games of varying difficulty where the goal is to collect a coin located at the end of a chain of rooms. In contrast to previous text-based RL approaches, we observe that our agent learns policies that generalize to unseen games of greater difficulty.

\end{abstract}

\section{Introduction}
\label{sect:intro}
  Text-based games like Zork~\citep{if:Zork} are complex, interactive simulations.
  They use natural language to describe the state of the world, to accept actions from the player, and to report subsequent changes in the environment. The player works toward goals which are seldom specified explicitly and must be discovered through exploration.
  The observation and action spaces in text games are both combinatorial and compositional, and players must contend with partial observability, since descriptive text does not communicate complete, unambiguous information about the underlying game state.
  
  In this paper, we study several methods of exploration in text-based games. Our basic task is a deterministic text-based version of the \textit{chain experiment}~\citep{osband2016,plappert2017parameter} with distractor nodes that are off-chain: the agent must navigate a path composed of discrete locations (rooms) to the goal, ideally without revisiting dead ends.
  We propose a DQN-based recurrent model for solving text-based games, where the recurrence gives the model the capacity to condition its policy on historical state information.
  To encourage exploration, we extend count-based exploration approaches~\citep{ostrovski2017count,tang2017exploration}, which assign an intrinsic reward derived from the count of state visitations during learning, across episodes.
  Specifically, we propose an episodic count-based exploration scheme, where state counts are reset at the beginning of each episode. This reward plays the role of an \emph{episodic memory}~\citep{gershman2017reinforcement} that pushes the agent to visit states not previously encountered within an episode.
  Although the recurrent policy architecture has the capacity to solve the task by remembering and avoiding previously visited locations, we hypothesize that exploration rewards will help the agent learn to utilize its memory.
  
  We generate a set of games of varying difficulty (measured with respect to the path length and the number of off-chain rooms) with a text-based game generator~\citep{textworld}. We observe that, in contrast to a baseline model and standard count-based exploration methods, the recurrent model with episodic bonus learns policies that not only complete multiple training games at same time successfully but also generalize to \emph{unseen} games of \emph{greater difficulty}.

\section{Text-based Games as POMDPs}
\label{sect:rl}
  Text-based games are sequential decision-making problems that can be described naturally by the Reinforcement Learning (RL) setting.
  Fundamentally, text-based games are partially observable Markov decision processes (POMDP)~\citep{kaelbling1998planning} where the environment state is never observed directly. To act optimally, an agent must keep track of all observations. 
  Formally, a text-based game is a discrete-time POMDP defined by $(S, T, A, \Omega, O, R, \gamma)$, where $\gamma \in [0, 1]$ is the discount factor.

  \textbf{Environment States ($S$):}
  The environment state at turn $t$ in the game is $s_t \in S$. It contains the complete internal information of the game, much of which is hidden from the agent. When an agent issues a command $\bc_t$ (defined next), the environment transitions to state $s_{t+1}$ with probability $T(s_{t+1} | s_t, \bc_t)$.

  \textbf{Actions ($A$):}
  At each turn $t$, the agent issues a text command $\bc_t$. The interpreter can accept any sequence of characters but will only recognize a tiny subset thereof. Furthermore, only a fraction of recognized commands will actually change the state of the world. The resulting action space is enormous and intractable for existing RL algorithms. In this work, we make the following two simplifying assumptions. (1) \textbf{Word-level} Each command is a two-word sequence where the words are taken from a fixed vocabulary $\vocab$. (2) \textbf{Command syntax} Each command is a $\mathtt{(verb, object)}$ pair (direction words are considered objects).
  
  \textbf{Observations ($\Omega$):}
  The text information perceived by the agent at a given turn $t$ in the game is the agent's observation, $o_t \in \Omega$, which depends on the environment state and the previous command with probability $O(o_t|s_t,\bc_{t-1})$. Thus, the function $O$ selects from the environment state what information to show to the agent given the last command.

  \textbf{Reward Function ($R$):}
  Based on its actions, the agent receives reward signals $r_t = R(s_t, a_t)$. The goal is to maximize the expected discounted sum of rewards $E \left[\sum_t \gamma^t r_t \right]$.

\section{Method}
  \label{sect:model}
  \subsection{Model Architecture}
      In this work, we adopt the LSTM-DQN~\citep{narasimhan2015} model as baseline. It has two modules: a representation generator $\Phi_R$, and an action scorer $\Phi_A$.
      $\Phi_R$ takes observation strings $o$ as input, after a stacked embedding layer and LSTM~\citep{hochreiter1997lstm} encoder, a mean-pooling layer produces a vector representation of the observation. This feeds into $\Phi_A$, in which two MLPs, sharing a lower layer, predict the Q-values over all verbs $w_v$ and object words $w_o$ independently. The average of the two resulting scores gives the Q-values for the composed actions. The LSTM-DQN does not condition on previous actions or observations, so it cannot deal with partial observability. We concatenate the previous command $\bc_{t-1}$ to the current observation $o_t$ to lessen this limitation.

\begin{figure}
      \centering
      \includegraphics[width=0.45\textwidth]{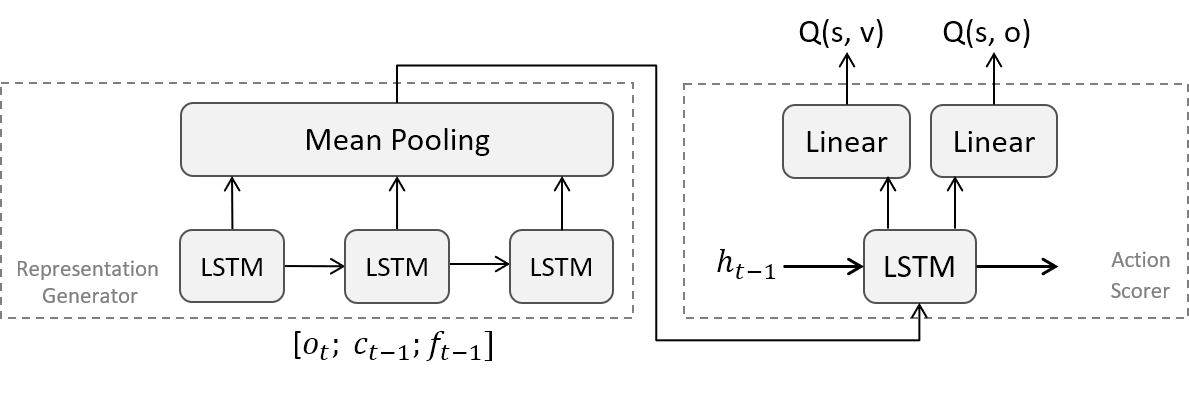}
      \caption{\textbf{LSTM-DRQN} processes textual observations word-by-word to generate a fixed-length vector representation. This representation is used by the recurrent policy to estimate Q-values for all verbs $Q(s,v)$ and objects $Q(s,o)$.}
      \vspace{-1em}
      \label{fig:model}
      \end{figure}

      To further enhance the agent's capacity to remember previous states, we replace the shared MLP in $\Phi_A$ by an LSTM cell. This model is inspired by \cite{hausknecht2015drqn, lample2016fps} and we call it LSTM-DRQN. The LSTM cell in $\Phi_A$ takes the representation generated by $\Phi_R$ together with history information $h_{t-1}$ from the previous game step as input. It generates the state information at the current game step, which is then fed into the two MLPs as well as passed forward to next game step. Figure~\ref{fig:model} shows the LSTM-DRQN architecture.
      
  \subsection{Discovery Bonus}
      To promote exploration we use an intrinsic reward by counting state visits~\citep{kolter2009near,tang2017exploration,martin2017count,ostrovski2017count}. We investigate two approaches to counting rewards. The first is inspired by \cite{kolter2009near}, where we define the \textbf{cumulative counting bonus} as $r^{+}(o_t) = \beta \cdot n(o_t)^{-1/3}$,
      where $n(o_t)$ is the number of times the agent has observed $o_t$ since the beginning of training (across episodes), and $\beta$ is the bonus coefficient. During training, as the agent observes new states more and more, the cumulative counting bonus gradually converges to 0.
      
      The second approach is the \textbf{episodic discovery bonus}, which encourages the agent to discover unseen states by assigning a positive reward whenever it sees a new state. It is defined as: 
      $ \footnotesize
        r^{++}(o_t) = 
            \begin{cases}
                \beta & \text{if $n(o_t) = 1$}\\
                0.0 & \text{otherwise}
            \end{cases}
      $, where $n(\cdot)$ is reset to zero at the beginning of each episode. Taking inspiration from \cite{gershman2017reinforcement}, we hope this behavior pushes the agent to visit states not previously encountered in the current episode and teaches the agent how to use its memory for this purpose so it may generalize to unseen environments.
  
\section{Related Work}
  \label{sect:related}
  \textbf{RL Applied to Text-based Games:}
  \citet{narasimhan2015} test their LSTM-DQN in two text-based environments: Home World and Fantasy World. They report the quest completion ratio over multiple runs but not how many steps it takes to complete them.
  \citet{he2015deep} introduce the Deep Reinforcement Relevance Network (DRRN) for tackling choice-based (as opposed to parser-based) text games, evaluating the DRRN on one deterministic game and one larger-scale stochastic game. 
  The DRRN model converges on both games; however, this model must know in advance the valid commands at each state.
  \citet{fulda2017can} propose a method to reduce the action space for parser-based games by training word embeddings to be aware of verb-noun affordances. 
  One drawback of this approach is it requires pre-trained embeddings.
  
  \textbf{Count-based Exploration:}
  The Model Based Interval Estimation-Exploration Bonus (MBIE-EB)~\cite{strehl2008analysis} derives an intrinsic reward by counting state-action pairs with a table $n(s, a)$. Their exploration bonus has the form $\beta / \sqrt{n(s,a)}$ to encourage exploring less-visited pairs.
  In this work, we use $n(s)$ rather than $n(s,a)$, since the majority of actions leave the agent in the same state (\ie unrecognized commands). Using the latter would reward the agent for trying invalid commands, which is not sensible in our setting.
 
  \citet{tang2017exploration} propose a hashing function for count-based exploration in order to discretize high-dimensional, continuous state spaces. Their exploration bonus $r^{+} = \beta / \sqrt{n(\phi(s))}$,
  where $\phi(\cdot)$ is a hashing function that can either be static or learned. This is similar to the \textit{cumulative counting bonus} defined above.
  
  \textbf{Deep Recurrent Q-Learning:}
  \citet{hausknecht2015drqn} propose the Deep Recurrent Q-Networks (DRQN), adding a recurrent neural network (such as an LSTM~\citep{hochreiter1997lstm}) on top of the standard DQN model. DRQN estimates $Q(o_t, h_{t-1}, a_t)$ instead of $Q(o_t, a_t)$, so it has the capacity to memorize the state history. \citet{lample2016fps} use a model built on the DRQN architecture to learn to play FPS games.
  
  A major difference between the work presented in this paper and the related work is that we test on unseen games and train on a set of similar (but not identical) games rather than training and testing on the same game.

\section{Experiments}
  \label{sect:experiments}
  
  \begin{figure*}
      \centering
      \includegraphics[width=0.72\textwidth]{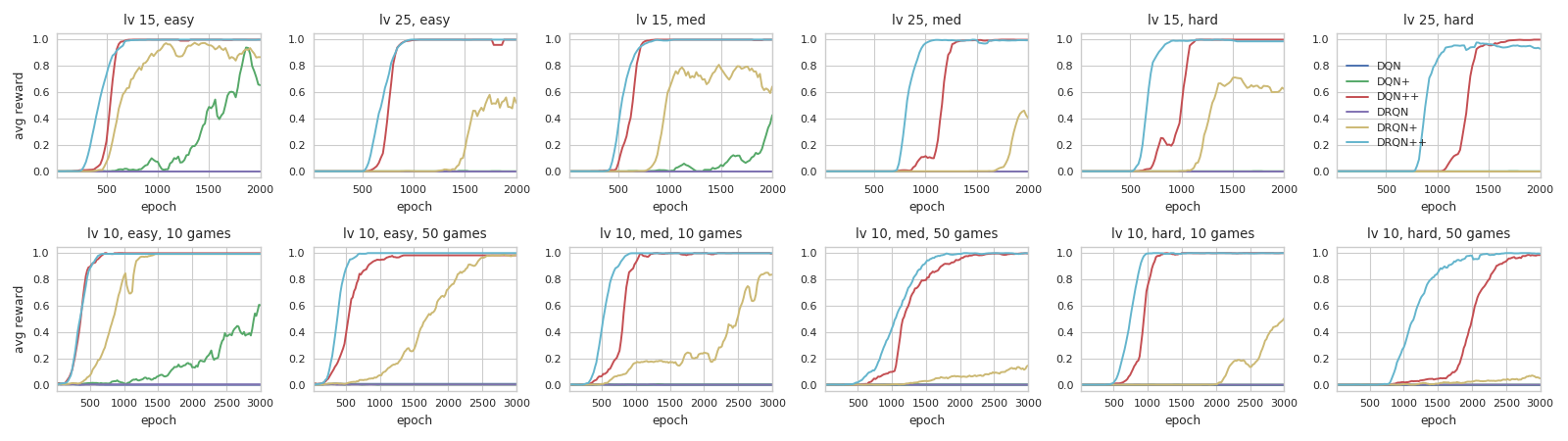}
      \caption{Model performance on single games (top row) and multiple games (bottom row).}
      \label{fig:single_multi}
  \end{figure*}
  
  \subsection{Coin Collector Game Setup}
  To evaluate the two models described above and the proposed discovery bonus, we designed a set of simple text-based games inspired by the chain experiment~\citep{osband2016,plappert2017parameter}. Each game contains a given number of rooms that are randomly connected to each other to form a chain (see figures in Appendix~\ref{sec:example}). The goal is to find and collect a \literal{coin} placed in one of the rooms. The player's initial position is at one end of the chain and the coin is at the other. These games have deterministic state transitions.
  
  Games stop after a set number of steps or after the player has collected the coin. The game interpreter understands only five commands (\cmd{go north}, \cmd{go east}, \cmd{go south}, \cmd{go west} and \cmd{take coin}), while the action space is twice as large: \{\cmd{go}, \cmd{take}\} $\times$ \{\cmd{north}, \cmd{south}, \cmd{east}, \cmd{west}, \cmd{coin}\}. See Figure~\ref{fig:coin_collector_text}, Appendix~\ref{sec:example} for an example of what the agent observes in-game.
  
  Our games have 3 \textbf{modes}: \emph{easy} (mode 0), there are no distractor rooms (dead ends) along the path; \emph{medium} (mode 1), each room along the optimal trajectory has one distractor room randomly connected to it; \emph{hard} (mode 2), each room on the path has two distractor rooms, \ie within a room on the optimal trajectory, all 4 directions lead to a connected room. We use difficulty \textbf{levels} to indicate the optimal trajectory's length of a game.
  
  To solve easy games, the agent must learn to recall its previous directional action and to issue the command that does not reverse it (\eg if the agent entered the current room by going \cmd{east}, do not now \cmd{go west}). Conversely, to solve medium and hard games, the agent must reverse its previous action when it enters distractor rooms to return to the chain, and also recall farther into the past to track which exits it has already passed through. Alternatively, since there are no cycles, it can learn a less memory intensive ``wall-following'' strategy by, \eg taking exits in a clockwise order from where it enters a room.
    
  We refer to models with the cumulative counting bonus as $\var{model+}$, and models with episodic discovery bonus as $\var{model++}$, where $\var{model} \in \{\var{DQN}, \var{DRQN}\}$\footnote{Since all models use the LSTM representation generator, we omit ``LSTM'' for abbreviation.} (implementation details in Appendix~\ref{sec:implement_detail}). In this section we cover part of the experiment results, the full extent of our experiment results are provided in Appendix~\ref{sec:more_results}.
 
  \subsection{Solving Training Games}
  
  We first investigate whether the variant models can learn to solve single games with different difficulty modes (easy, medium, hard) and levels $\{L5, L10, L15, L20, L25, L30\}$\footnote{We use $Lk$ to indicate level $k$ game.}. As shown in Figure~\ref{fig:single_multi} (top row), when the games are simple, vanilla DQN and DRQN already fail to learn. Adding the cumulative bonus helps somewhat and models perform similarly with and without recurrence. When the games become harder, the cumulative bonus helps less, while episodic bonus remains very helpful and recurrence in the model becomes very helpful.
  
  Next, we are interested to see whether models can learn to solve a distribution of games. Note that each game has its own counting memory, \ie the states visited in one game do not affect the counters for other games. Here, we fix the game difficulty level to 10, and randomly generate training sets that contain $\{2, 5, 10, 30, 50, 100\}$ games in each mode. As shown in Figure~\ref{fig:single_multi} (bottom row), when the game mode becomes harder, the episodic bonus has an advantage over the cumulative bonus, and recurrence becomes more crucial for memorizing the game distribution. It is also clear that the episodic bonus and recurrence help significantly when more training games are provided.
  
  \subsection{Zero-shot Evaluation}
  \begin{figure*}
  \centering
  \includegraphics[width=0.8\textwidth]{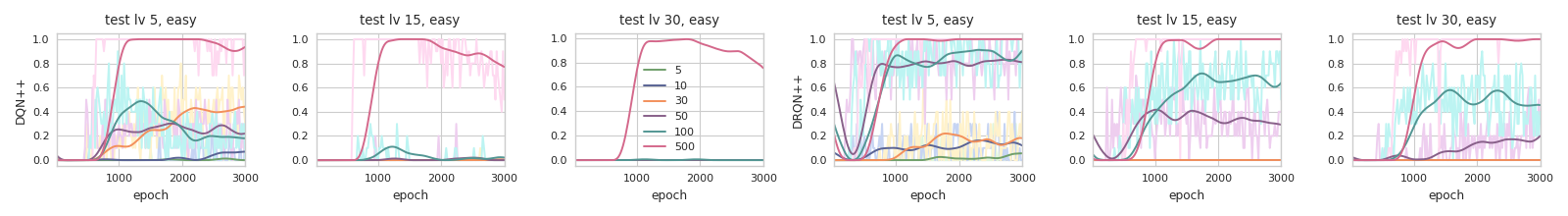}
  \caption{Zero-shot evaluation: Average rewards of DQN++ (left) and DRQN++ (right) as a function of the number of games in the training set.}
  \vspace{-1em}
  \label{fig:zero_shot}
  \end{figure*}
  
  Finally, we want to see if a pre-trained model can generalize to unseen games. The generated training set contains $\{1, 2, 5, 10, 30, 50, 100, 500\}$ L10 games for each mode. Then, for each corresponding mode the test set contains 10 \emph{unseen} $\{L5, L10, L15, L20, L30\}$ games. There is no overlap between training and test games in either text descriptions or optimal trajectories. At test time, the counting modules are disabled, the agent is not updated, and its generates verb and noun actions based on the $argmax$ of their Q-values.
  
  As shown in Figure~\ref{fig:zero_shot}, when the game mode is easy, both models with and without recurrence can generalize well on unseen games by training on a large training set. It is worth noting that by training on 500 L10 easy games, both models can almost perfectly solve level 30 unseen easy games. We also observe that models with recurrence are able to generalize better when trained on fewer games.
  
  \begin{figure}
  \centering
  \includegraphics[width=0.48\textwidth]{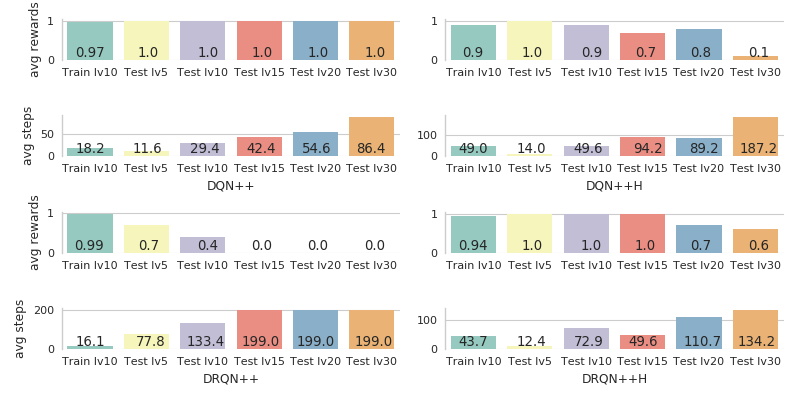}
  \caption{Average rewards and steps used corresponding to best validation performance in hard games. }
  \label{fig:zero_shot_hard}
  \end{figure}
  
  When testing on hard mode games, we observe that both models suffer from overfitting (after a certain number of episodes, average test reward starts to decrease while training reward increases). Therefore, we further generated a validation set that contains 10 L10 hard games, and report test results corresponding to best validation performance. In addition, we investigated what happens when concatenating the previous 4 steps' history observation into the input. In Figure~\ref{fig:zero_shot_hard}, we add $H$ to model names to indicate this variant.
  
  As shown in Figure~\ref{fig:zero_shot_hard}, all models can memorize the 500 training games, while $\var{DQN++}$ and $\var{DRQN++H}$ are able to generalize better on unseen games. In particular, the former performs near perfectly on test games. To investigate this, we looked into all the bi-grams of generated commands (\ie two commands from adjacent game steps) from $\var{DQN++}$ model. Surprisingly, except for moving back from dead end rooms, the agent always explores exits in anti-clockwise order. This means the agent has learned a general strategy that does not require history information beyond the previous command. This strategy generalizes perfectly to all possible hard games because there are no cycles in the maps. 
  
\section{Final Remarks}
  We propose an RL model with a recurrent component, together with an episodic count-based exploration scheme that promotes the agent's discovery of the game environment. 
  We show promising results on a set of generated text-based games of varying difficulty. In contrast to baselines, our approach learns policies that generalize to unseen games of greater difficulty.
  
  In future work, we plan to experiment on games with more complex topology, such as cycles (where the ``wall-following'' strategy will not work). We would like to explore games that require multi-word commands (\eg unlock red door with red key), necessitating a model that generates sequences of words. Other interesting directions include agents that learn to map or to deal with stochastic transitions in text-based games.


\clearpage
\bibliography{main}
\bibliographystyle{icml2018}

\clearpage

\appendix
\section{Implementation Details}
\label{sec:implement_detail}
Implementation details of our neural baseline agent are as follows\footnote{Our implementation is publicly available at \url{https://github.com/xingdi-eric-yuan/TextWorld-Coin-Collector}.}.
In all experiments, the word embeddings are initialized with 20-dimensional random matrices; the number of hidden units of the encoder LSTM is 100. In the non-recurrent action scorer we use a 1-layer MLP which has 64 hidden units, with $ReLU$ as non-linear activation function, in the recurrent action scorer, we use an LSTM cell which hidden size is 64.

In replay memory, we used a memory with capacity of $500000$, a mini-batch gradient update is performed every 4 steps in the gameplay, the mini-batch size is 32. We apply prioritized sampling in all experiments, in which, we used $\rho=0.25$. In LSTM-DQN and LSTM-DRQN model, we used discount factor $\gamma=0.9$, in all models with discovery bonus, we used $\gamma=0.5$.

When updating models with recurrent components, we follow the update strategy in \cite{lample2016fps}, \ie we randomly sample sequences of length 8 from the replay memory, zero initialize hidden state and cell state, use the first 4 states to bootstrap a reliable hidden state and cell state, and then update on rest of the sequence.

We anneal the $\epsilon$ for $\epsilon$-greedy from 1 to 0.2 over 1000 epochs, it remains at 0.2 afterwards. In both cumulative and episodic discovery bonus, we use coefficient $\beta$ of 1.0. 

When zero-shot evaluating hard games, we use $max\_train\_step = 100$, in all other experiments we use $max\_train\_step = 50$; during test, we always use $max\_test\_step = 200$.

We use \emph{adam} \citep{kingma2014adam} as the step rule for optimization. The learning rate is $1e^{-3}$. The model is implemented using \textit{PyTorch} \citep{paszke2017automatic}.

All games are generated using TextWorld framework~\citep{textworld}, we used the house grammar.

\clearpage
\section{More Results}
\label{sec:more_results}

  \begin{figure}[h!]
  \begin{center}
  \includegraphics[width=0.9\textwidth]{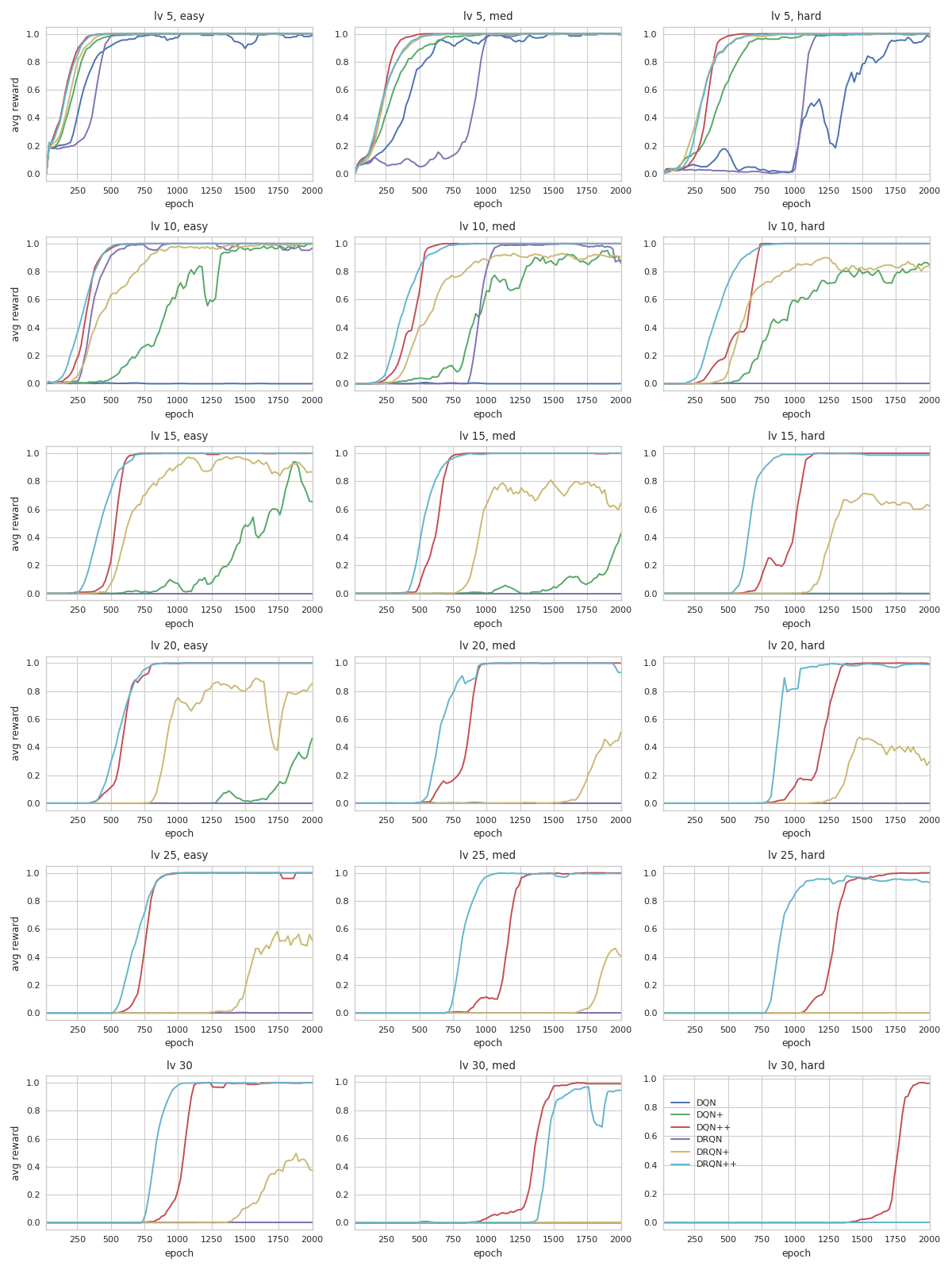}
  \caption{Model performance on single games.}
  \label{fig:more_single}
  \end{center}
  \end{figure}
  
  \begin{figure*}
  \begin{center}
  \includegraphics[width=0.9\textwidth]{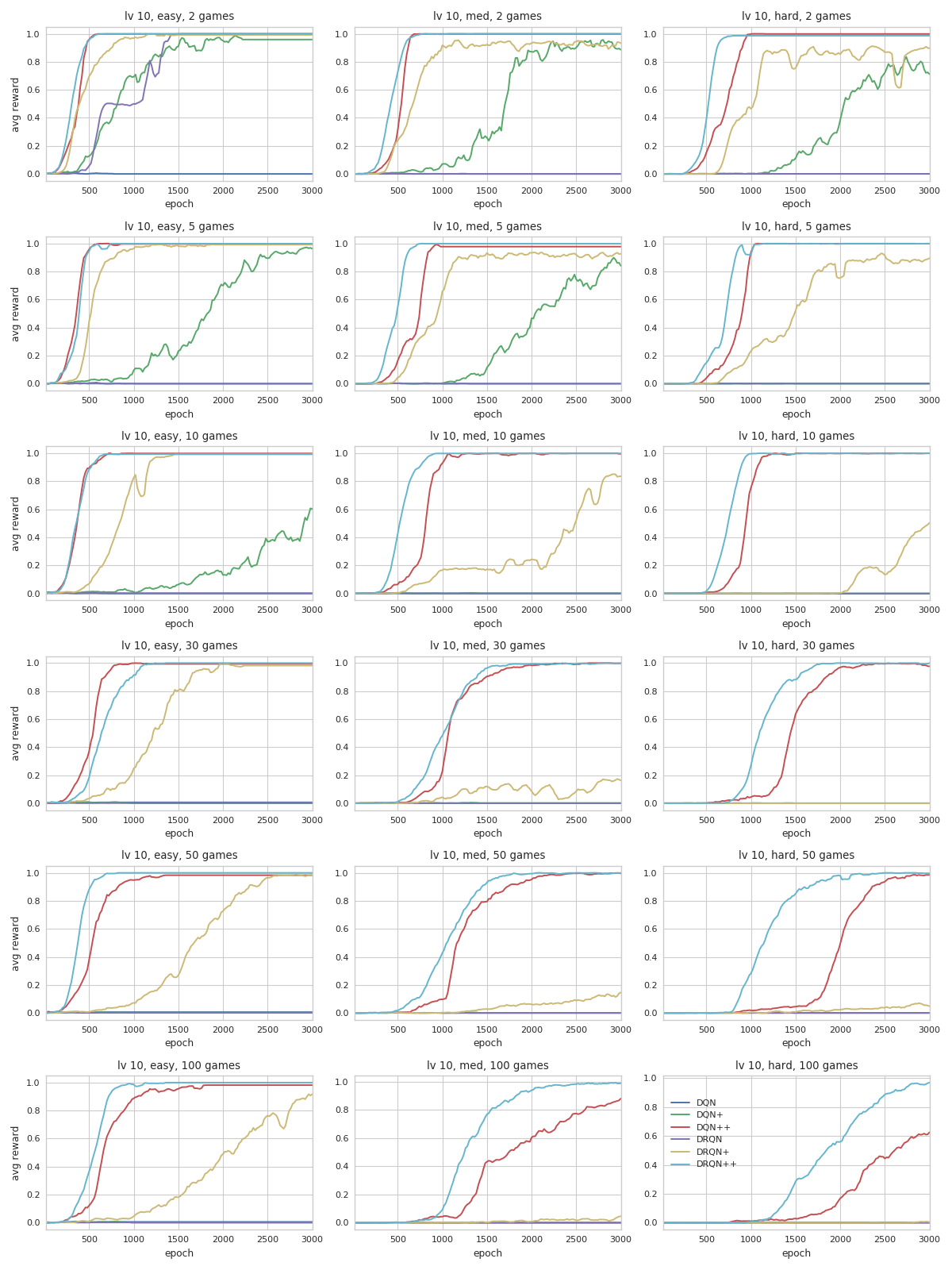}
  \caption{Model performance on multiple games.}
  \label{fig:more_multi}
  \end{center}
  \end{figure*}
  
  \begin{figure*}
  \begin{center}
  \includegraphics[width=0.7\textwidth]{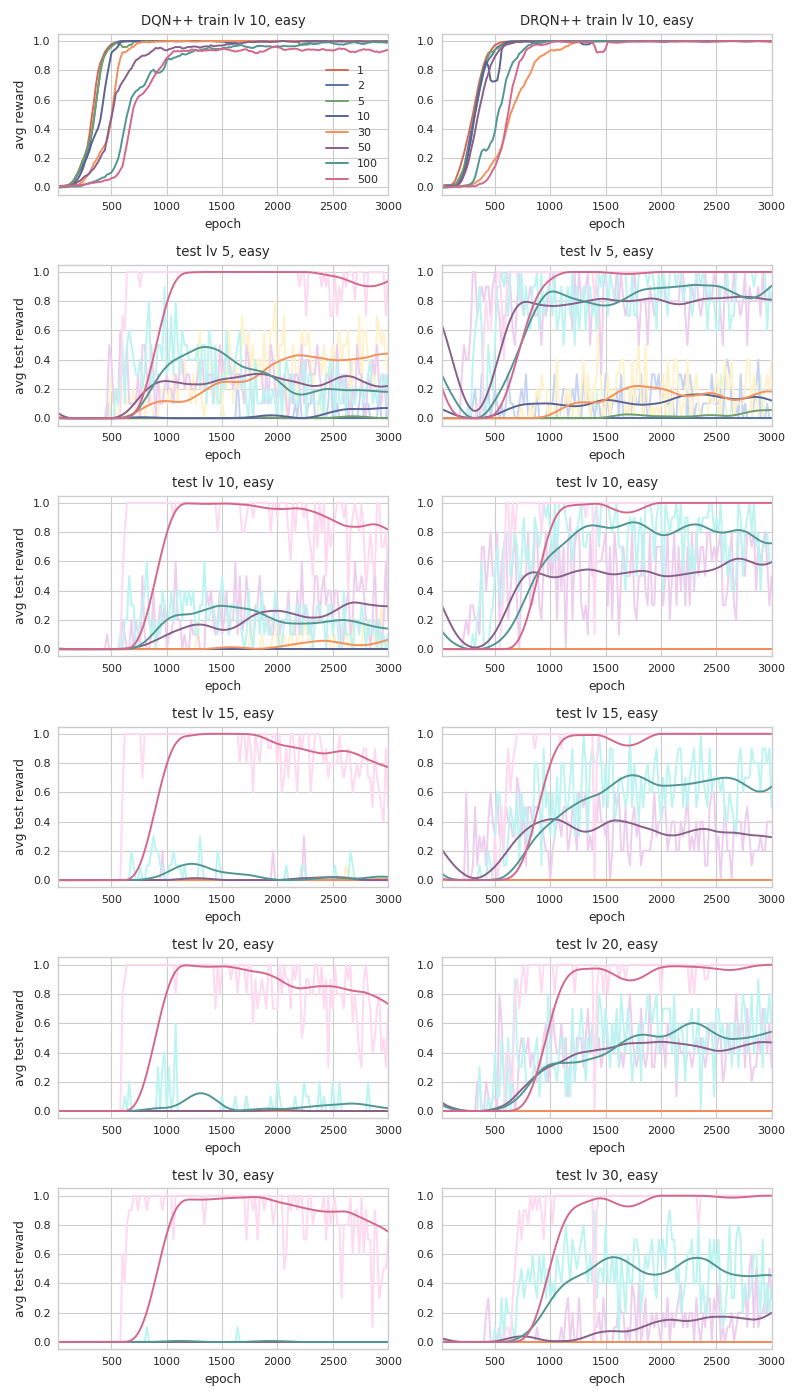}
  \caption{Model performance on unseen easy test games when pre-trained on easy games.}
  \label{fig:more_test_easy}
  \end{center}
  \end{figure*}
  
  \begin{figure*}
  \begin{center}
  \includegraphics[width=0.7\textwidth]{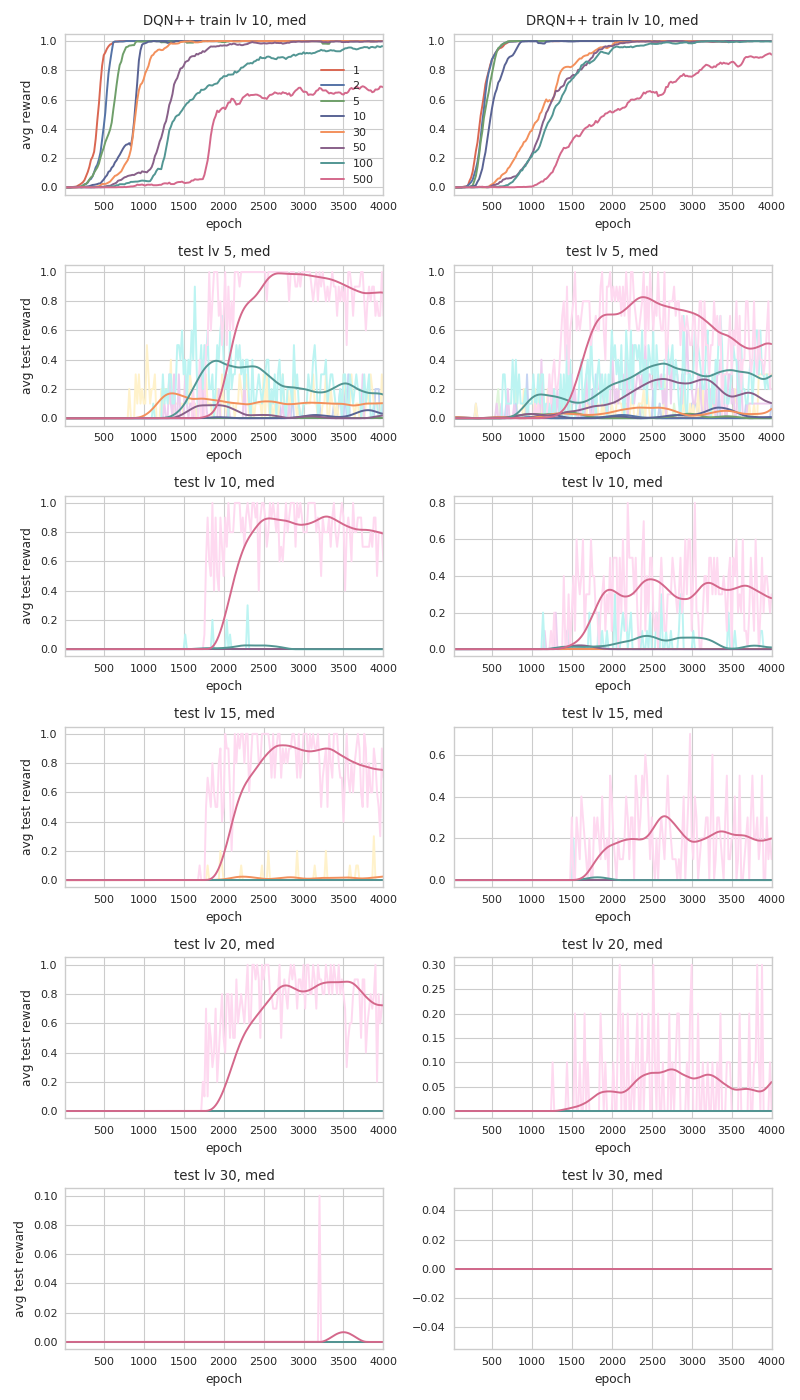}
  \caption{Model performance on unseen medium test games when pre-trained on medium games.}
  \label{fig:more_test_med}
  \end{center}
  \end{figure*}
  
\clearpage
\onecolumn
\section{Text-based Chain Experiment}
\label{sec:example}
  \begin{figure*}[h!]
  \centering
      \quad\quad\quad\quad  
      \includegraphics[width=0.3\textwidth]{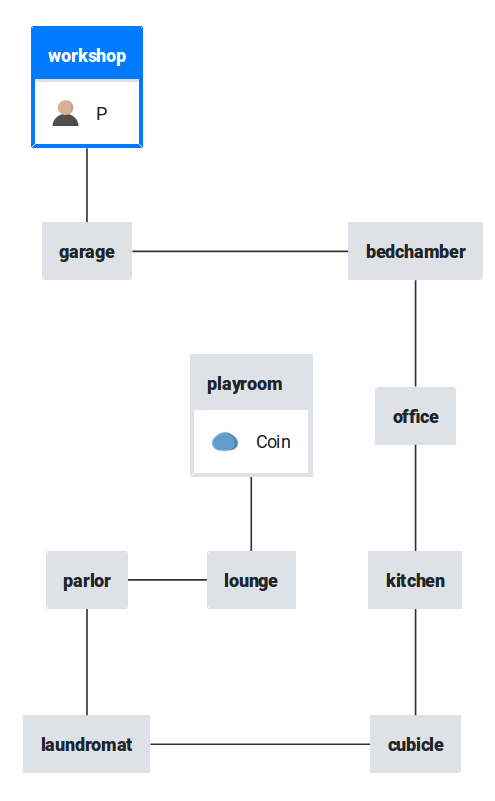}
  \caption{Examples of the games used in the experiments: level 10, easy}
  \label{fig:coin_collector_easy}
  \end{figure*}
  
  \begin{figure*}[h!]
  \centering
      \quad\quad\quad\quad  
      \includegraphics[width=0.7\textwidth]{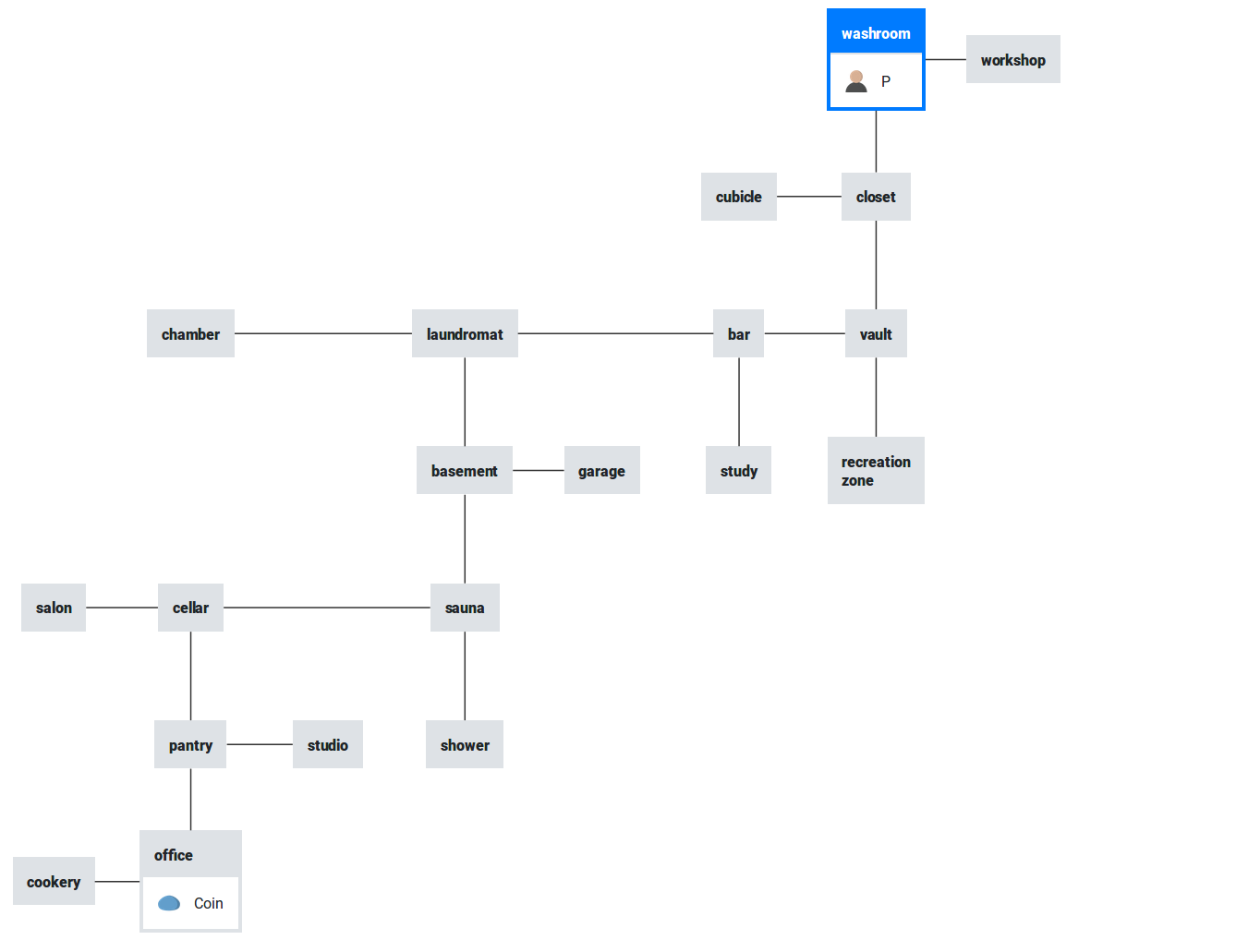}
  \caption{Examples of the games used in the experiments: level 10, medium}
  \label{fig:coin_collector_med}
  \end{figure*}
  
  \begin{figure*}
  \centering
      \quad\quad\quad\quad  
      \includegraphics[width=0.8\textwidth]{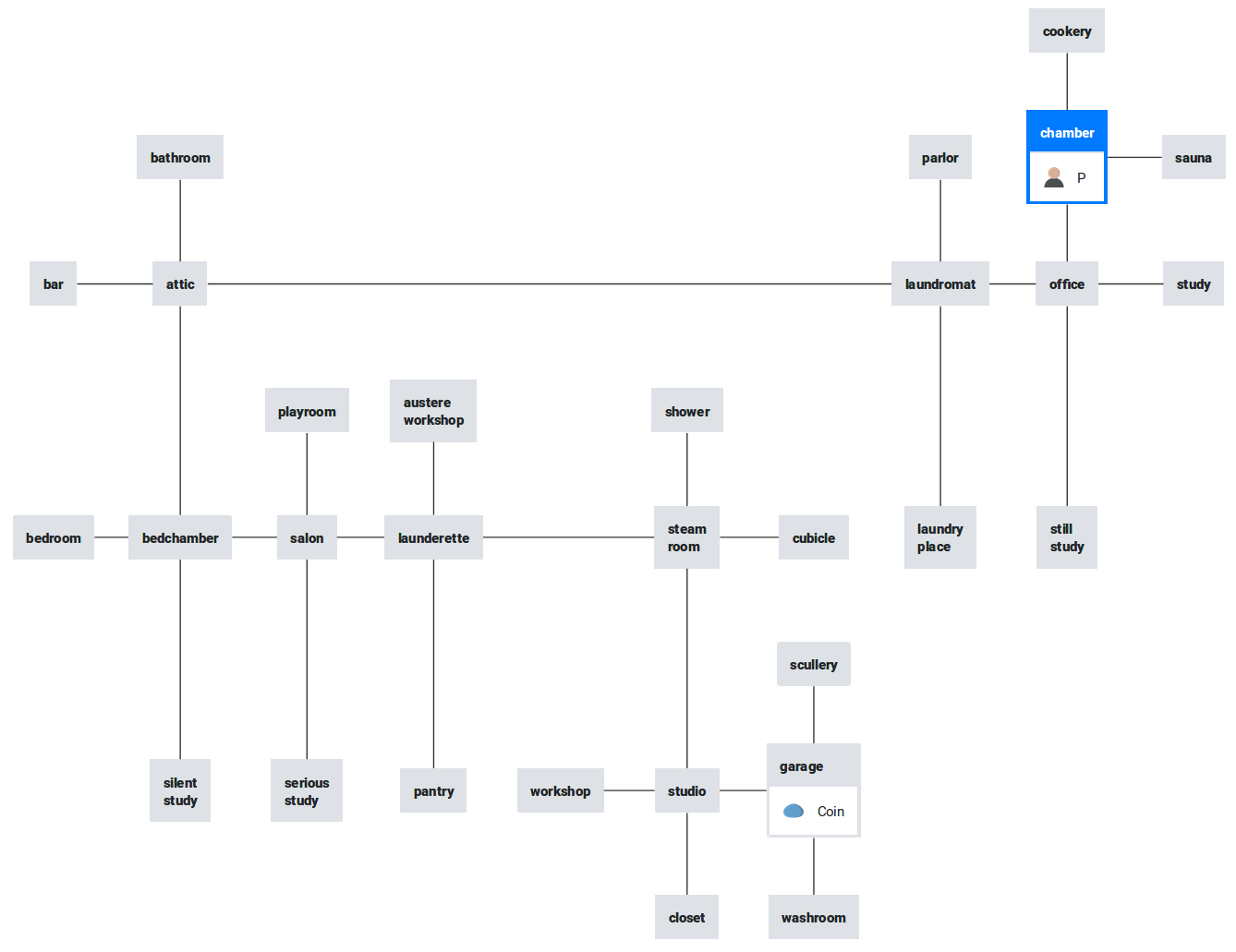}
  \caption{Examples of the games used in the experiments: level 10, hard}
  \label{fig:coin_collector_hard}
  \end{figure*}
  
  \begin{figure*}
  \begin{center}
    \includegraphics[width=0.8\textwidth]{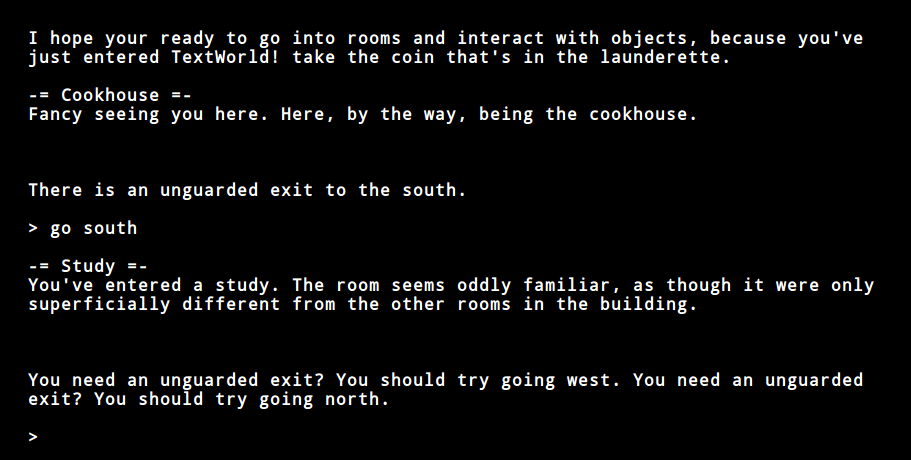}
  \caption{Text the agent gets to observe for one of the level 10 easy games.}
  \label{fig:coin_collector_text}
  \end{center}
  \end{figure*}

\end{document}